\documentclass[runningheads]{llncs}
\usepackage{graphicx}
\usepackage{amsmath,amssymb} 
\usepackage{color}
\usepackage{url,bbm,booktabs,array,algorithm,algpseudocode}
\usepackage{cite}
\usepackage{multirow, makecell, array}
\begin{document}
\pagestyle{headings}
\mainmatter

\def\ACCV20SubNumber{127}  

\title{Domain Adaptation Gaze Estimation by Embedding with Prediction Consistency} 
\titlerunning{Domain Adaptation Gaze Estimation}
%
\author{Zidong Guo\inst{1} \and
Zejian Yuan\inst{1} \and
Chong Zhang\inst{2} \and
Wanchao Chi\inst{2} \and
Yonggen Ling\inst{2} \and
Shenghao Zhang\inst{2}
}
%
\authorrunning{Guo et al.}
%
\institute{Institute of Artificial Intelligence and Robotics, Xi’an Jiaotong University, China \\
\email{gzd3118311122@stu.xjtu.edu.cn, yuan.ze.jian@xjtu.edu.cn} \and
Tencent Robotics X, China \\
\email{aerentzhang@gmail.com, wanchaochi@tencent.com, ylingaa@connect.ust.hk, popshzhang@pku.edu.cn}
}


\maketitle

\begin{abstract}
Gaze is the essential manifestation of human attention. In recent years, a series of work has achieved high accuracy in gaze estimation. However, the inter-personal difference limits the reduction of the subject-independent gaze estimation error. This paper proposes an unsupervised method for domain adaptation gaze estimation to eliminate the impact of inter-personal diversity. In domain adaption, we design an embedding representation with prediction consistency to ensure that linear relationships between gaze directions in different domains remain consistent on gaze space and embedding space. Specifically, we employ source gaze to form a locally linear representation in the gaze space for each target domain prediction. Then the same linear combinations are applied in the embedding space to generate hypothesis embedding for the target domain sample, remaining prediction consistency. The deviation between the target and source domain is reduced by approximating the predicted and hypothesis embedding for the target domain sample. Guided by the proposed strategy, we design Domain Adaptation Gaze Estimation Network(DAGEN), which learns embedding with prediction consistency and achieves state-of-the-art results on both the MPIIGaze and the EYEDIAP datasets.
\end{abstract}

\section{Introduction}

Gaze servers as an important visual cue of human attention. Accurate gaze estimation can provide critical support for many applications, such as human-computer interaction \cite{hci}, virtual reality \cite{VR}, and driver monitoring systems \cite{DMS}. Although eye tracker can provide a precise gaze estimation \cite{tracking}, the high price and the demand for specific equipment limit its applications in the real world and more flexible environments. Unconstrained appearance-based gaze estimation methods can predict 2D gaze target position or 3D gaze angles based on patches cropped from RGB images. Thanks to the advancement of convolutional neural networks (CNN) and a large number of publicly available high-quality datasets, the error of gaze estimation has been dramatically decreased in recent years.

\begin{figure}[t]
\centering
\includegraphics[width=120mm]{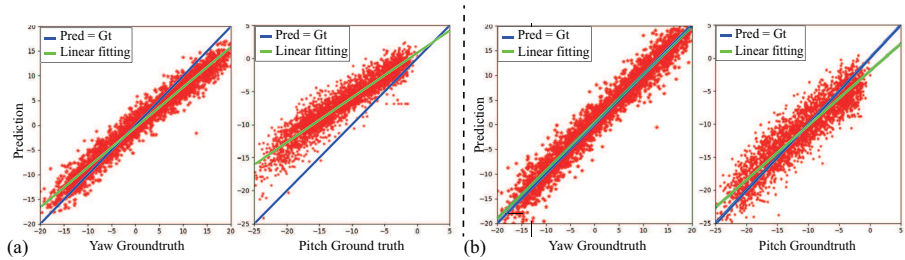} 
\caption{
Scatter plot of the groundtruth (X-axis) and the network gaze estimation (Y-axis) of the yaw and pitch angles in an evaluation set from the MPIIGaze dataset. The results are estimated by (a) Regression from eye region based on CNN and (b) DAGEN (Ours).
}
\label{fig:question}
\end{figure}

Appearance-based gaze estimation can decouple gaze direction from high-dimension images with various noises, but some challenges still restrict the further improvement of estimation precision. Obtaining gaze groundtruth requires specific equipment, a well-defined collection strategy, and highly concentrated attention of participants \cite{utmultiview,eyediap, mpiigaze, RT-GENE, gaze360, mpiigaze_pami}. Under these strict conditions, current datasets violate the identical independent distribution(i.i.d) nature, that is, only tens of persons participating in the collection of thousands of gaze direction data per subject \cite{menet}. For gaze estimation that requires open set testing [12], the deviation between the distribution of the training set and the test set is reflected in the prediction as a person-specific bias. As shown in Fig. \ref{fig:question}(a), the bias between the network regression and the groundtruth can often be observed, which is also mentioned in [13].  Some methods perform a person-specific gaze estimation through several new subject's labeled data to eliminate the bias \cite{diffnet, FAZE}. However, in practice, even a bit of accurately labeled data is challenging to acquire.

In this work, we propose an unsupervised method for domain adaptation (DA) gaze estimation to eliminate the impact of inter-personal differences and fit new subject's data without groundtruth labels. In domain adaption, we design an embedding representation with prediction consistency to ensure that linear relationships between gaze directions remain consistent on gaze space and embedding space. We then build the Domain Adaptation Gaze Estimation Network (DAGEN) using EPC loss devised to measure this consistency. Moreover, a new training strategy is employed for domain adaptation.

The most crucial element of DAGEN is the embedding with prediction consistency (EPC), which is expected to eliminate the deviation between domains. Following the Locally Linear Embedding (LLE) representation method \cite{lle}, the hypothesis label predicted on the target domain is linearly interpreted by its neighbor gaze directions from the source domain. Such linear combinations in gaze space would be migrated to embedding space to obtain hypothesis embedding, which ensures locally linear consistency between the embedding and prediction space. For the same gaze directions, we demand the embedding features encoding gaze should also be similar. However, due to the deviation between domains, the embedding also retains some domain-specific features unrelated to gaze direction, which causes a fixed bias in the test. So EPC loss, which weighs the distance between target hypothesis embedding and predicted embedding for each target domain sample, can be used to illustrate the deviation between domains. We optimize the EPC loss to eliminate the deviation between domains, thus achieving domain adaptation. We present our DAGEN estimation results in Fig. \ref{fig:question}(b) to exhibit the consequence of domain adaptation. 

We evaluate our proposed method on two commonly used gaze datasets and indicate that our DAGEN can effectively improve the accuracy of gaze estimation. On both datasets, our estimation results exceeding the current state-of-the-art method. Specifically, the DAGEN achieves a 9.66$\%$ improvement (4.14$^{\circ}$ to 3.74$^{\circ}$) on MPIIGaze, and an 18.9$\%$ improvement (5.3$^{\circ}$ to 4.3$^{\circ}$) on EYEDIAP. Note that the input only uses the eye region patch, and the source and target domain are the train and evaluation set, respectively.

The major contributions of our work are summarized as follows:

1). We propose a new representation for the target domain embedding with prediction consistency, as a linear combination of neighbors from the source domain.

2). We design an innovative embedding with prediction consistency (EPC) loss for unsupervised domain adaptation gaze estimation, enabling it to measure the shift between the source and target domain. 

3). Our method achieves state-of-the-art performance on MPIIGaze and EYEDIAP with only eye region as input.

\section{Related Work}
Gaze estimation methods are typically divided into appearance-based and model-based methods \cite{gazemethod}. Model-based methods rely on the biological structure and reflection characteristics of the eyeball, and usually require high-resolution images with homogeneous illumination \cite{model1, model2}. Appearance-based methods can robustly decouple gaze angles from high-dimensional images with various noises. Recently, due to the application of many large data sets \cite{mpiigaze, mpiigaze_pami, eyediap, iTracker} and the development of CNNs, the accuracy of appearance-based gaze estimation methods has been continuously improved. 
Zhang et al. first use LeNet \cite{lenet} structure based on CNN and MPIIGaze dataset to process gaze estimation\cite{mpiigaze}. Subsequently, many works have improved the accuracy of gaze estimation through different methods. For example, multi-modal input was utilized in \cite{iTracker}; the key role of face was proved in \cite{SWCNN}; a new convolution paradigm especially for gaze estimation was devised in \cite{delatednet}; timing information was used in \cite{RCNN}; the four models ensemble method was used to increase estimation accuracy in \cite{RT-GENE}; and a coarse-to-fine estimation strategy was designed in \cite{aaai} .

However, recent work has discovered the fixed deviation in gaze estimation caused by person-specific diversity, as shown in Fig. \ref{fig:question}(a). 
The diversity is reduced by learning gaze differences and applying calibration sets in \cite{diffnet}. Random effects, which actively learns the differences among-subjects during training, was introduced in \cite{menet}. And a meta-learning method, performing person-specific calibration for each new subject and generating a person-specific network, is utilized in \cite{FAZE} to eliminate deviations.

Domain adaptation improves prediction performance in the target domain by aligning the distribution from the source domain \cite{da1, da2}. Some work attempts to minimize the discrepancy between domains to obtain domain-invariant features directly \cite{mmd, jmmd}. Recently, some methods found that aligning targets in both domains could significantly increase prediction performance. For instance, \cite{cannet} uses the correlation between classes to perform domain adaptation by predicting the target hypothesis label and source groundtruth. For the gaze estimation problem, discriminator is applied to distinguish the source and target domains, thereby aligning the domains\cite{gaze360}. The differences was taken advantage in pairs of gaze directions and performs domain adaptation through gaze redirection and cycle consistency \cite{cycleconsis}. 

However, these methods do not address the deviations caused by subject-difference. We propose an unsupervised domain adaptation method to eliminate the inter-personal differences by introducing embedding with prediction consistency.

\section{Proposed Method}
\label{method}
%
%
%
%
Domain Adaptation (DA) is applied to solve the inter-personal differences reflected in the domain shift in the data distribution. Our method takes an eye region image $I$ as input, regressing $g=(y, p)$ through a feature extractor and a linear mapping, where $y$ and $p$ means yaw and pitch in gaze direction. Given a source domain $S=\{(I_{1}^{s}, g_{1}^{s}), \cdots , (I_{N_{s}}^{s}, g_{N_{s}}^{s})\}$ with several participants and groundtruth, and a target domain $T=\{I_{1}^{t}, \cdots , I_{N_{t}}^{t}\}$ using test set data without groundtruth, the proposed network adopts Domain Adaptation (DA) as the training strategy to align the embedding between $S$ and $T$ to increase its estimation performance. 

\begin{figure}[t]
\centering
\includegraphics[width=100mm]{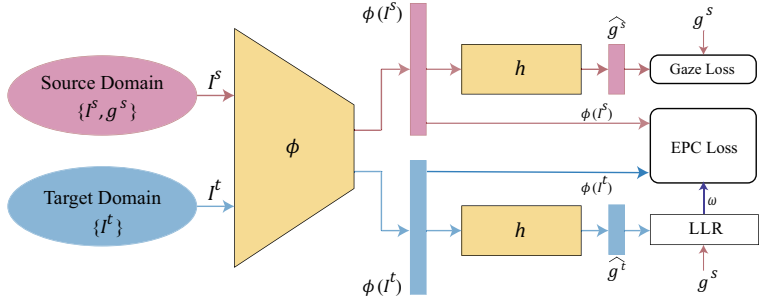} 
\caption{
The architecture of our proposed DAGEN. ResNet-18 and a fully connected layer are employed as feature exactor $\phi$. A linear mapping $h$ maps embedding feature $\phi(I)$ to gaze prediction $\hat{g}$. During DA training, LLR is utilized to generate linear weight $w$ by source groundtruth and target prediction. Besides gaze loss for source domain, we apply EPC loss for unsupervised learning embedding features. 
}
\label{net}
\end{figure}

Figure 2 provides an architecture of our Domain Adaptation Gaze Estimation Network (DAGEN). The feature extractor $\phi(\cdot)$ contains an ImageNet \cite{Imagenet} pre-trained ResNet-18 \cite{resnet} followed a multilayer perceptron. The embedding feature $\phi(I)$ will be constrained to keep consistency with predicted gaze direction during DA training. Finally, as the restrained prediction consistency embedding could decouple gaze-related information from the high-dimension image, the gaze direction $\hat{g}$ is calculated through a simple linear mapping operation $h$.

\subsection{Target Domain Gaze Representation}
\label{sec:LLG}
We propose a Locally Linear Representation (LLR) for Target Domain Gaze that employs source domain gaze to represent the target hypothesis label in gaze space $\mathbb{G}$ linearly. For each sample in the target domain, the network prediction is considered as a hypothesis label. We linearly combine $k$ source domain samples in its neighborhood in the $\mathbb{G}$ to describe it.

We first define the neighborhood for each target domain prediction in $\mathbb{G}$ to ensure the correct representation. Only when both angles in target hypothesis label $\hat{g_{j}^{t}}$ and source gaze direction $g_{i}^{s}$ are not much different (less than $\mu$), would $g_{i}^{s}$ be set as a neighborhood of $\hat{g_{j}^{t}}$. We describe the set of all neighbors of $\hat{g_{j}^{t}}$ as $\mathcal{N}_j$, defined as,

\begin{equation}
 \mathcal{N}_j = \left\{g_{i}^{s} | \max \;(|y_{i}^{s}-\hat{y_{j}^{t}}|,\; |p_{i}^{s}-\hat{p_{j}^{t}}|) < \mu \right\}.
\label{select}
\end{equation}

Every target domain prediction $\hat{g_{j}^{t}}$ in a mini-batch having over $k$ neighbors would be randomly selected $k$ neighbors to regenerate $\mathcal{N}_j$, which is employed to reconstruct the $\hat{g_{j}^{t}}$. We define the weight $w_{ij}$ to summarize the contribution of the $i$th data in $\mathcal{N}_{j}$ to the $\hat{g_{j}^{t}}$ reconstruction, and the purpose is to find a suitable solution of each $w_{ij}$. 

For 2D gaze direction $g$, the slightly larger number of neighbors means that it is challenging to find a suitable solution to minimize the reconstruction loss $E\left(w\right)$ during training. We consider involving more neighbors in the reconstruction of $\hat{g_{j}^{t}}$ and introduce an L2 regularization term to ensure a unique solution. So an L2 regularization term is suitable to solve this problem. The reconstruction loss $E\left(w\right)$ is formally expressed as, 

\begin{equation}
 \begin{aligned}
 &E\left(W_{j}\right) = \|{\hat{g_{j}^{t}} - \sum _{i=1}^{k}}w_{ji}g_{i}^{s}\|_{2}^{2} + \lambda\sum _{i=1}^{k}w_{ji}^{2},&s.t.\ g_{i}^{s} \in \mathcal{N}_j\ and\ \sum _{i=1}^{k}w_{ji}=1,
 \end{aligned}
\label{L2-LLE}
\end{equation}

where $W_{j}=[w_{j1},\cdots,w_{jk}]$. The Eq. (\ref{L2-LLE}) can be written in matrix form as, 
\begin{equation}
 \begin{aligned}
 E \left(W_{j}\right) &= W_{j}^{T} (\hat{G_{j}^{t}} - G_{i}^{s})^{T} (\hat{G_{j}^{t}} - G_{i}^{s}) W_{j} + \lambda W_{j}^{T}W_{j},\\
 &= W_{j}^{T}(S_{j} + \lambda I)W_{j},
 \end{aligned}
\label{L2-LLE-matrix}
\end{equation}

\begin{figure}[t]
\centering
\includegraphics[width=100mm]{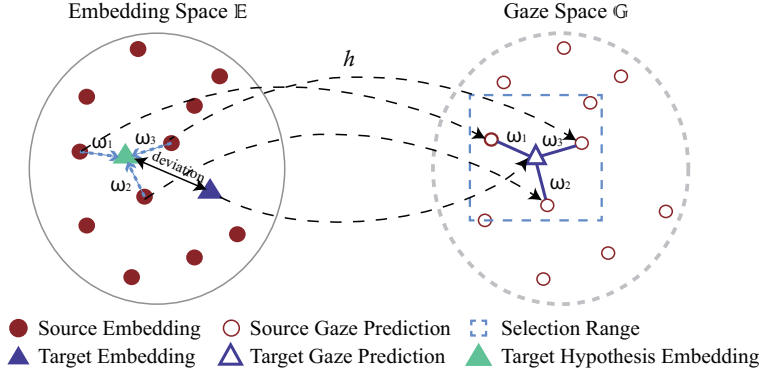} 
\caption{
Embedding with prediction consistency. The linear combination relationship in $\mathbb{G}$ is inherited to $\mathbb{E}$ to generate a hypothesis embedding $\hat{\phi}(I^{t})$ for each target sample. The distance between $\hat{\phi}(I^{t})$ and the target predicted embedding $\phi(I^{t})$ measures the deviation between the source and target domain.
}
\label{fig:consis}
\end{figure}

where $\hat{G_{j}^{t}}=[\hat{g_{j}^{t}},\cdots,\hat{g_{j}^{t}}]_{1\times k}$, $G_{i}^{s}=[g_{1}^{s},\cdots,g_{k}^{s}]$, and $S_{j}$ is regarded as a local covariance matrix, defined as,

\begin{equation}
    S_{j}=(\hat{G_{j}^{t}} - G_{i}^{s})^{T} (\hat{G_{j}^{t}} - G_{i}^{s}).
\end{equation}

The solution $W^{*}_{j}$ that minimizes $E\left(W_{j}\right)$ obtained by the Lagrange multiplier method is,

\begin{equation}
 W^{*}_{j}=\frac{(S_{j} + \lambda I)^{-1}1_{k}}{1_{k}^{T}(S_{j} + \lambda I)^{-1}1_{k}}.
\label{solution}
\end{equation}

With the optimized weight $W^{*}_{j}=[w^{*}_{j1},\cdots,w^{*}_{jk}]$, LLR is formally described as,

\begin{equation}
 \hat{g_{j}^{t}}=\sum _{i=1}^{k}w^{*}_{ji}g_{i}^{s}, \ \ \ g_{i}^{s} \in \mathcal{N}_j.
\label{emb}
\end{equation}

\subsection{Embedding with Prediction Consistency}

Here we propose Embedding with Prediction Consistency (EPC) for domain adaptation. EPC transfers the same linear combination relationship in gaze space $\mathbb{G}$ to embedding space $\mathbb{E}$ to generate target hypothesis embeddings. For target domain sample $I^{t}_{j}$, the hypothesis embedding is declared as, 
\begin{equation}
 \hat{\phi}(I_{j}^{t})=\sum _{i=1}^{k}w^{*}_{ji}\phi(I_{i}^{s}), \ \ \ g_{i}^{s} \in \mathcal{N}_j.
\label{emb}
\end{equation}

As shown in Fig. \ref{fig:consis}, the LLR weight in gaze space $\mathbb{G}$ are inherited to the embedding space $\mathbb{E}$. For each target predicted embedding in $\mathbb{E}$, we generate the target hypothesis embedding $\hat{\phi}(I^{t})$ by Eq. (\ref{emb}). The linear relationship between target hypothesis embedding and source predicted embedding is the same as that between target and source gaze directions, which is the embedding with prediction consistency.


\subsection{Loss Function}
With the purpose of domain adaptation between the source and target domain, we propose DA loss consisting of two items, as shown in Eq. (\ref{total}). Specifically, $L_{EPC}$ measures the deviation between the source and target domain. Meanwhile, $L_{gaze}$ supervises the predicted gaze directions of the source domain to guarantee that the network is always optimized towards reducing gaze estimation error. 

\begin{equation}
 L_{DA}=\lambda_{EPC}L_{EPC}+\lambda_{gaze}L_{gaze},
\label{total}
\end{equation}
where we empirically set$\lambda_{EPC}=1$ and $\lambda_{gaze}=1$. 

We introduce an embedding with prediction consistency (EPC) loss for domain adaptation gaze estimation, which ensures same gaze directions should have the same embedding features unrelated to any interferences like appearance. Typically this constraint requires pairs of images in totally same gaze directions from different subjects. However, it is nearly unreachable to meet this condition for continuous gaze direction. As mentioned in section \ref{sec:LLG}, for each target gaze hypothesis label, we employ LLR of adjacent source gaze to indicate it. The combination relationships in $\mathbb{G}$ are transferred to $\mathbb{E}$ to generate target hypothesis embedding remaining prediction consistency. 

Given a batch of $B_{s}$ source image samples and $B_{t}$ target image samples during training, we formally compute the $L_{EPC}$ using,

\begin{equation}
 L_{EPC} = \frac{1}{B_{t}} \sum_{j=1}^{B_{t}} d (\phi \left(I_{j}^{t}\right), \;\sum _{i=1}^{k}w^{*}_{ji}\phi(I_{i}^{s})), \ \ \ h(\phi(I_{i}^{s})) \in \mathcal{N}_j,
\label{consistency}
\end{equation}
where $L1$ distance is employed as the function $d$. $L_{EPC}$ measures the distance between the hypothesis and predicted embedding. Furthermore, since target hypothesis embedding is a linear combination of source predicted embedding, $L_{EPC}$ also evaluates the deviation between the source and target domains. During training, as target hypothesis embedding and predicted embedding get closer and closer, the offset between domains is gradually eliminated.

Besides preserving $L_{EPC}$ for domain adaptation, the source domain with groundtruth should also take part in parameter updating to guide training optimizing. The $L_{gaze}$ is calculated based on cosine similarity as, 

\begin{equation}
 L_{gaze}\left(\hat{g^{s}},\; g^{s} \right) = \arccos{\frac{\hat{g^{s}} \times g^{s}}{\left\|\hat{g^{s}}\right \| \cdot \left\|{g^{s}}\right \|}}.
\label{gaze_loss}
\end{equation}

\subsection{Training}
Since the network prediction decides the neighborhood and locally linear gaze representation in $L_{EPC}$, a well-trained model is necessary to generate credible target hypothesis labels. We pre-train the network using only source domain with groundtruth and the $L_{gaze}$ for $N_{s}$ epochs at first.

\renewcommand{\algorithmicrequire}{\textbf{Input:}}
\renewcommand{\algorithmicensure}{\textbf{Output:}}
\algnewcommand{\algorithmicgoto}{\textbf{Output:}}
\algnewcommand{\Goto}[1]{\algorithmicgoto~\ref{#1}}
\begin{algorithm}[t]
\caption{Training Procedure}
\label{training}
\begin{algorithmic}[0]
\Require \\
 Source Domain: $S=\{(I_{1}^{s}, g_{1}^{s}), \cdots , (I_{N_{s}}^{s}, g_{N_{s}}^{s})\}$ \\
 Target Domain: $T=\{I_{1}^{t}, \cdots , I_{N_{t}}^{t}\}$
\Ensure \\
Model parameter $\theta ^{*}$
\end{algorithmic}
\begin{algorithmic}[1]
\State \textcolor{blue}{\textbf{\# First Step: Pre-training in the Source Domain}}
\For{$m \mbox{ in } \left[1, N_{s}\right]$}
 \For{$(I_{i}^{s}, g_{i}^{s}) \mbox{ in } S$}
 \State Forward $I_{i}^{s}$ and obtain prediction $\hat{g_{i}^{s}}$.
 \State Back-propagation with Eq.(\ref{gaze_loss}) and update network parameters $\theta$.
 \EndFor
\EndFor
\State \textcolor{blue}{\textbf{\# Second Step: Joint Optimization}}
\For{$m \mbox{ in } \left[1, M_{t}\right]$}
 \State Sample a mini-batch $B_{s}$ and $B_{t}$ from $S$ and $T$.
 \State Obtain prediction $\hat{g^{s}}$ and $\hat{g^{t}}$ with forwarding $I^{s}$ and $I^{s}$.
 \For{$b \mbox{ in } [1, B_{t}]$}
 \State Select qualified sample set $\mathcal{N}_{b}$ from $B_{s}$ for $\hat{g_{t}^{b}}$. (Eq. (\ref{select}))
 \If {$\|\mathcal{N}_{b}\| \textless $ k}
 \State Continue
 \Else
 \State Randomly choosing $k$ samples from $\mathcal{N}_{b}$.
 \State Obtain LLR representation $W^{*}$ of $\hat{g_{b}}$ using $k$ samples (Eq.(\ref{solution}))
 \State Calculate hypothesis embedding $\hat{\phi}(I^{t}_{b})$ by Eq.(\ref{emb})
 \State Compute $L_{EPC}$ using Eq.(\ref{consistency})
 \EndIf
 \State Compute $L_{gaze}$ for $B_{s}$ using Eq.(\ref{gaze_loss})
 \State Back-propagation with Eq.(\ref{total}) and update network parameters $\theta$.
 \EndFor
\EndFor
\end{algorithmic}
\end{algorithm}

In the joint training procedure, we need to optimize the $L_{gaze}$ and $L_{EPC}$ simultaneously for $M_{t}$ iterations. We employ an alternative optimization strategy \cite{cannet} to perform each iteration. Specifically, we first update target hypothesis labels $\hat{g^{t}}$ with network parameters fixed in each loop and meanwhile estimate the prediction of the source domain. Then given the target label $\hat{g^{t}}$, we construct $\mathcal{N}_{j}$ and estimate $L_{EPC}$. It is worth mentioning that we use the source domain groundtruth for LLR to obtain higher estimation accuracy. Furthermore, network parameters are updated by back-propagation to minimize $L_{EPC}$ and $L_{gaze}$ finally.

Algorithm \ref{training} summarizes the entire optimization precessing of our DAGEN. First step performs the essential pre-training, and Second step shows the joint optimization procedure. We asynchronously update the target label and optimize the network to ensure the effectiveness and efficiency during training. We use SGD with momentum = $0.9$ as the optimizer and a base learning rate of $0.001$, l2 weight regularization of $5 \times 10^{-4}$. $B_{s}$ and $B_{t}$ are both set to $64$ during training.

\section{Experiments}
\subsection{Datasets}
We implement the proposed Domain Adaptation Gaze Estimation Network on two current gaze datasets: MPIIGaze \cite{mpiigaze} and EYEDIAP \cite{eyediap}.

\subsubsection{MPIIGaze} is a very challenging dataset for appearance-based in-the-wild gaze direction estimation, because it has high within-subject variations in facial appearance and environments, for instance, make-up, hair change, illumination intensity and direction. We only use the standard evaluation subset MPIIFaceGaze provided by MPIIGaze, which contains 37667 images captured from 15 subjects and has facial keypoints label for image pre-processing. 

\subsubsection{EYEDIAP} contains 94 video sequences of 16 subjects, who were looking at screen targets or physical targets in the collection. Only the videos collected with screen target sessions are used in our training and evaluation set. Note that, since two participants lack the videos in the screen target session, we sample one image every fifteen frames from the other 14 subjects. 

\begin{figure}[t]
\centering
\includegraphics[width=120mm]{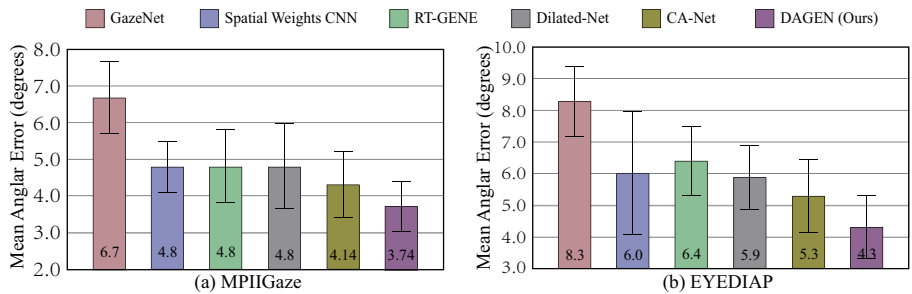} 
\caption{
Performance of gaze estimation on (a)MPIIGaze and (b)EYEDIAP using a leave-one-subject-out strategy. Bars represent the MAE, and the specific value in degrees is on the bottom of each bar; error bars indicate standard deviations.
}
\label{fig:compare}
\end{figure}

\subsection{Data Pre-processing}
We manipulate the pre-processing procedure similar to \cite{preprocess, SWCNN, mpiigaze_pami} to normalize two datasets, and utilize the Surrey Face Model as the reference 3D face model. In the appearance-based gaze estimation task, the head pose has a significant influence on the accuracy since its six freedoms bring calculational complexity and time-consuming. Consequently, we select four eye corners and two mouth corners described in \cite{mpiigaze_pami} for PnP-based head pose estimation. Then transfer and rotate the virtual camera according to the head pose to eliminate the impact of position and roll angle. 

In our work, considering applying a single image as input cover both eyes, we select the mean of four 3D eye corner landmarks as the gaze origin point to produce groundtruth for source domain. We normalize the camera's intrinsic parameters with a focal length of $960\; \mbox{mm}$, and a distance of $410\;\mbox{mm}$ from the face to generate image patches of size $256 \times 64$ as input for training. In each test period, in order to better verify the effect of domain adaptation, we use the newcomer's entire data without groundtruth as the target domain.

\subsection{Comparison with Appearance-Based Methods}
\setlength{\tabcolsep}{4pt}
\begin{table}[t]
\begin{center}
\caption{
Comparison of Appearance-Based Gaze Estimation Methods. 
}
\label{compare}
\begin{tabular}{cccccc}
\toprule
\makebox[10mm]{Methods} & 
\makebox[18mm]{Input} & 
\makebox[8mm]{Data} & 
\makebox[8mm]{GT} & 
\makebox[14mm]{MPIIGaze} & 
\makebox[14mm]{EYEDIAP}\\
\midrule
GazeNet \cite{mpiigaze} & left eye + head pose &$\times$ & $\times$ & 6.7$^{\circ}$ & 8.3$^{\circ}$\\
\noalign{\smallskip}
\noalign{\smallskip}
SWCNN \cite{SWCNN} & face &$\times$ & $\times$ & 4.8$^{\circ}$ & 6.0$^{\circ}$\\
\noalign{\smallskip}
\noalign{\smallskip}
RT-GENE \cite{RT-GENE} & two eyes + face &$\times$ & $\times$ & 4.8$^{\circ}$ & 6.4$^{\circ}$\\
\noalign{\smallskip}
\noalign{\smallskip}
Dilated-Net \cite{delatednet} & two eyes + face &$\times$ & $\times$ & 4.8$^{\circ}$ & 5.9$^{\circ}$\\
\noalign{\smallskip}
\noalign{\smallskip}
MeNet \cite{menet} & face &$\times$ & $\times$ & 4.9$^{\circ}$ & --- \\
\noalign{\smallskip}
\noalign{\smallskip}
CA-Net \cite{aaai} & two eyes + face &$\times$ & $\times$ & 4.14$^{\circ}$ & 5.3$^{\circ}$\\
\midrule
FAZE (3-shot) \cite{FAZE} & eye area &$\checkmark$ & $\checkmark$ & 4.1$^{\circ}$ & --- \\
\noalign{\smallskip}
\noalign{\smallskip}
FAZE (256-shot) \cite{FAZE} & eye area &$\checkmark$ & $\checkmark$ & 3.75$^{\circ}$ & --- \\
\midrule
DAGEN (ours) & eye area &$\checkmark$ & $\times$ & {\bf 3.74$^{\circ}$} & {\bf 4.30$^{\circ}$}\\
\bottomrule
\end{tabular}
\end{center}
\end{table}
\setlength{\tabcolsep}{1.4pt}

We first compare the performance of the proposed method with the state-of-art appearance-based gaze estimation methods. The experiment is carried out in both MPIIGaze and EYEDIAP. For the evaluation protocol, we use leave-one-subject-out strategy on both MPIIGaze and EYEDIAP.

We choose several CNN-based methods proposed from 2015 to 2020 as comparisons, including GazeNet \cite{mpiigaze}, Spatial weights CNN (SWCNN) \cite{SWCNN}, RT-GENE \cite{RT-GENE}, Dilated-Net \cite{delatednet}, MeNet \cite{menet}, Faze \cite{FAZE} and CA-Net \cite{aaai}.

Although four models ensemble can increase the accuracy of RT-GENE, we do not show the result of that for fairness. Since initializing the model pre-trained on ImageNet can effectively improve accuracy, we apply this strategy for GazeNet, Spatial weights CNN and Dilated-Net refer to \cite{delatednet}. We only present the results in the author's paper for cases where source codes are not provided, or the evaluation protocol is different from us.

Fig. \ref{fig:compare}(a) shows the results of MPIIGaze. The Mean Angular Error (MAE) of most work in recent years has become about 4.8$^{\circ}$ without any person calibration. These methods all have characteristics, such as the use of multi-modal input, the introduction of attention mechanism, the implementation of new training methods, or a new convolution strategy suitable for gaze estimation. CA-Net more cleverly used the coarse-to-fine information from faces to eyes to achieve a breakthrough of about 0.66$^{\circ}$. Our method achieves 9.66$\%$ to 3.74$^{\circ}$ comparing to state-of-the-art method CA-Net with only eye area as input.

Fig. \ref{fig:compare}(b) shows the results in EYEDIAP. Due to the lower image resolution, the performance of EYEDIAP is generally worse than that of MPIIGaze. Many innovations in recent years still bring a significant breakthrough in performance, and the best accuracy obtained in \cite{aaai} has reached 5.3$^{\circ}$. We get an 18.9$\%$ increase with the state-of-the-art method to 4.3$^{\circ}$. 

Table \ref{compare} summarizes some differences and results of recent methods for reference, including that not illustrated in Fig. \ref{fig:compare}. The header \emph{Data} and \emph{GT} show whether the methods need data or groundtruth for a new subject before evaluation. It is noteworthy that some person-specific methods like few-shot (FAZE) have achieved a great improvement for gaze estimation. We show the result of FAZE\cite{FAZE} based on 3-shot and 256-shot within-MPIIGaze leave-one-person-out evaluation. With test images without labels, our method can obtain results close to 256-shot Faze, proving the effectiveness of domain adaptation.

\subsection{Ablation Study}
We further evaluate our method under different settings to better demonstrate the effectiveness of our various design choices in the DAGEN. For all ablation experiments, the source domain and test set's selection follows the leave-one-subject-out strategy on the MPIIGaze dataset. 

\subsubsection{Contribution of Domain Adaptation}
We first perform an ablation study to demonstrate the effect of domain adaptation. Specifically, we evaluate the consequence of adding domain adaptation, the impact of different target domain data, and the influence of domain adaptation objects. Table \ref{DAconfig} shows the experimental results and the only change is \emph{DA} is the choice of target domain.

\emph{Without DA} shows the baseline model supervised by the $L_{gaze}$ during training, having the MAE of 4.84$^{\circ}$. In order to better assess the impact of target domain data on accuracy, we compare the estimation accuracy using GazeCapture \cite{iTracker} as the target domain. For \emph{GC}, we randomly sample 20 images for each participant in GazeCapture. With a total of 1366 subjects and 27320 images, we get the MAE of 4.17$^{\circ}$. Moreover, we randomly select 100 participants to discuss the influence of diversity in the target domain, named \emph{GCsubset}. \emph{Eval} uses the evaluation set as the target domain. The results show that utilizing diverse and targeted samples as the target domain can effectively improve the estimation performance, which may have reference significance for practical application.

For our proposed method described in Section \ref{method}, we use the target hypothesis label and the source groundtruth as the domain adaptation targets. \emph{Pred} takes the source predicted value instead of groundtruth as the domain adaptive target, with the MAE of 3.99$^{\circ}$. Since in this case, errors in the source domain data would also affect the domain adaptation process. In other words, using groundtruth as the DA target produces more substantial constraints for the updating direction of the parameters.
\begin{table}[t]
\setlength{\abovecaptionskip}{-0.2cm}
\begin{center}
\caption{
Comparison on different DA configurations. 
}
\label{DAconfig}
\begin{tabular}{c|cccc}
\hline
\multirow{2}{*}{\makebox[20mm]{Without DA}} & 
\multicolumn{4}{c}{DA}\\
\cline{2-5}
& \makebox[22mm]{GCsubset} & 
\makebox[18mm]{GC} & 
\makebox[20mm]{Pred} &
\makebox[20mm]{Eval}\\
\hline
4.84$^{\circ}$ & 4.66$^{\circ}$ & 4.17$^{\circ}$ & 3.99$^{\circ}$ & {\bf 3.74$^{\circ}$}\\
\hline
\end{tabular}
\end{center}
\end{table}

\subsubsection{Effect of Feature Representation}
\begin{figure}[h]
\setlength{\belowdisplayskip}{2pt}
\centering
\includegraphics[width=100mm]{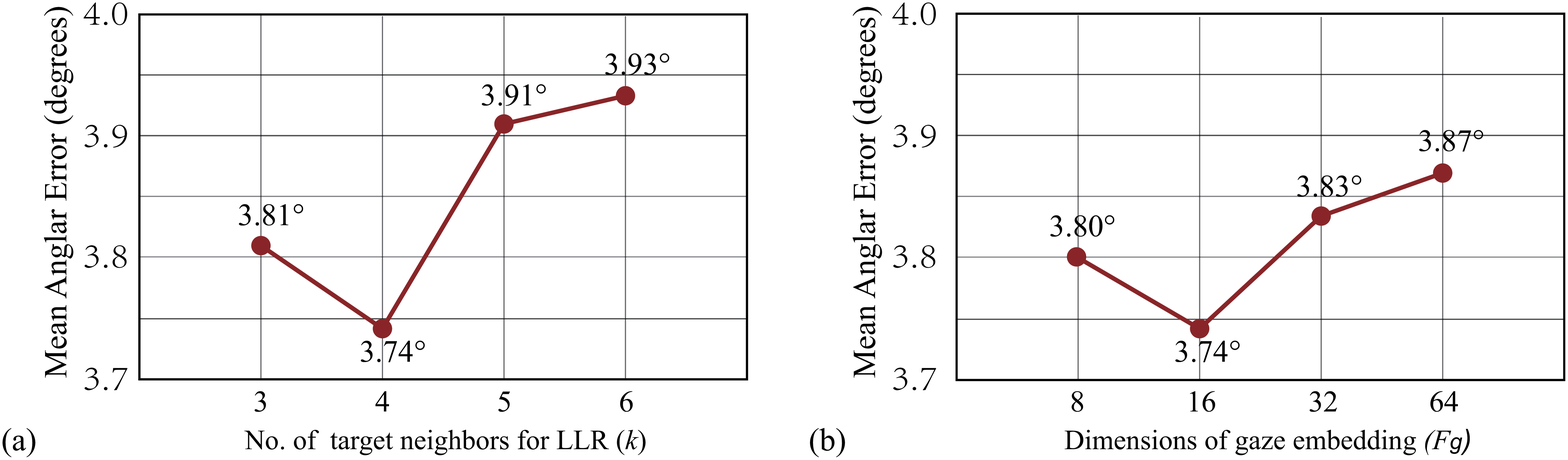} 
\caption{
Impact of different Feature Representation choice
}
\label{fig:represent}
\end{figure}

LLR utilizes $k$ source groundtruth to represent a target hypothesis label. We evaluate the different choices of $k$, shown in Fig. \ref{fig:represent}(a). Generally, a higher $k$ means more stable and robust LLR. However, because we select the appropriate sample from a mini-batch, a higher $k$ brings a smaller probability of reaching the selection condition. In our experimental protocol, the calculating speed of EPC loss is from $32.17 - 32.53 ms/iter$ in one Nvidia 1080Ti for different $k$, and the training speed is $76.5 ms/iter$ in training.

Our embedding $\phi \left(I\right)$ has the dimension of $F_{g}$. Considering $\phi(I)$ perform prediction consistency, different $F_{g}$ would lead to changes in characterization ability and robustness. We evaluate the accuracy of DAGEN for different dimensions $F_{g} = \{8, 16, 32, 64\}$ to select the most suitable one. Fig. \ref{fig:represent}(b) shows the result of dimension selection. In our experiments, our method is not sensitive to $F_{g}$, indicating that our method is very robust for $F_{g}$.

Empirically we find $k=4$ and $F_{g} = 16$ to be optimal hence select it.

\subsubsection{Effect of Pre-trained Model}
We use ResNet-18 pre-trained on ImageNet as the backbone. And before the domain adaptation training, the network is first trained for five epochs in the source domain. We evaluated the estimation accuracy of whether the two pre-trainings participate, shown in Table \ref{pretrain_compare}, to show the contribution of the two pre-training strategies.

For the model pre-trained on ImageNet,  it can effectively avoid the parameters falling into the local optimum, thereby improving the gaze estimation accuracy. In the case of pre-training in the source domain, obtaining a more accurate hypothesis label can significantly improve prediction accuracy. However, while the parameters fall into the local optimum, the quality of the hypothesis label is not improved, so the error is not substantially reduced.

\begin{table}[t]
\begin{minipage}{0.48\linewidth}
\centering
\caption{Impact of Pre-trained Methods.}
\label{pretrain_compare}
\begin{tabular}{ccc}
\toprule
\makebox[16mm]{ImageNet} & 
\makebox[16mm]{Source} &
\makebox[14mm]{MAE}\\
\midrule
$\times$ & $\times$ & 4.63$^{\circ}$\\
\noalign{\smallskip}
$\checkmark$ & $\times$ & 4.2$^{\circ}$\\
\noalign{\smallskip}
$\times$ & $\checkmark$ & 4.51$^{\circ}$\\
\noalign{\smallskip}
$\checkmark$ & $\checkmark$ & {\bf 3.74}$^{\circ}$\\
\bottomrule
\end{tabular}
\end{minipage}\begin{minipage}{0.55\linewidth}  
\centering
\caption{Impact of Selection Interval $\mu$.}
\label{dimension}
\begin{tabular}{cc}
\toprule
\makebox[20mm]{$\mu$} & 
\makebox[16mm]{MAE} \\
\midrule
0.05 & 4.04$^{\circ}$ \\
\noalign{\smallskip}
0.15 & {\bf 3.74$^{\circ}$} \\
\noalign{\smallskip}
0.3 & 3.84$^{\circ}$ \\
\noalign{\smallskip}
all & 3.86$^{\circ}$ \\
\bottomrule
\end{tabular} 
\end{minipage}
\end{table}
\subsubsection{Impact of Selection Range $\mu$}
The target hypothesis label needs to be represented by the appropriate source groundtruth. We have defined this selection strategy in Eq. (\ref{select}), where parameter $\mu$ indicates the select interval. We perform the impact of different choices of $\mu$ on estimation accuracy, shown in Table \ref{dimension}.

We can see that the estimation accuracy has not changed much when $\mu \geq 0.15$. The results reveal that although we established a locally linear relationship in the gaze space $\mathbb{G}$ and the embedding space $\mathbb{E}$, due to the linear mapping $h$, the network tends to exhibit a global linear relationship in $\mathbb{G}$ and $\mathbb{E}$. For the case where $\mu$ is small, few target samples can participate in DA training. Therefore, the network is straightforward to fall into overfitting, which significantly increases estimation error and even is challenging to converge.

\subsection{Visual Results}
\begin{figure}[t]
\centering
\includegraphics[width=110mm]{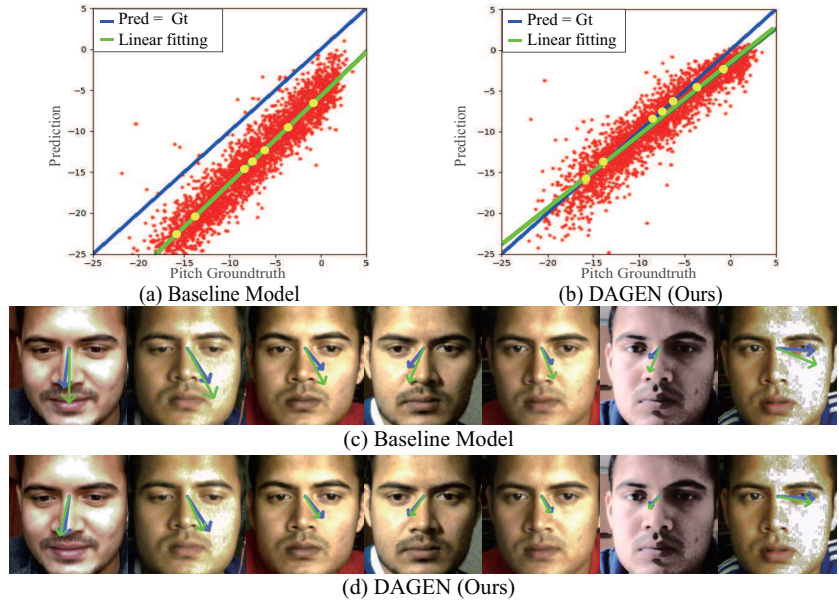} 
\caption{
Visible results of the evaluation set. (a) and (b) show the scatter (red) and linear fit line of pitch angles predicted with the baseline and our DAGEN. Yellow points in (a-b) are samples randomly selected from the linear fit line in (a). The groundtruth(blue) and prediction(green) of the chosen samples are displayed orderly in (c) and (d). 
}
\label{show}
\end{figure}

We display some results in Fig. \ref{show} to show the effectiveness of our method.  Fig. \ref{show}(a-b) performs the scatter plot and linear fit of the pitch angles, which are predicted by the baseline model and our proposed DAGEN method on the evaluation set. Obviously, the fixed bias between the prediction and the groundtruth is significantly reduced in our method. Furthermore, We randomly pick several samples close to the fitted line (yellow points in Fig \ref{show}(a-b) in the baseline results and visualize the result of both baseline and DAGEN models in Fig \ref{show}(c-d). We can see that our DAGEN can produce accurate gaze directions with tiny deviations for the evaluated subject in different appearances and illumination.

\section{Conclusion}
In this paper, we propose an unsupervised method for domain adaptation gaze estimation by embedding with prediction consistency.  We utilize source groundtruth to perform a locally linear representation for target gaze estimation.  The linear relationships are then inherited from gaze space to embedding space to perform prediction consistency. Moreover, we minimize the distance between the target hypothesis embedding and predicted embedding, which measures the deviation between the source and target domain. We experimentally showed that our approach dramatically reduces the impact of inter-personal differences and achieves state-of-the-art performance in MPIIGaze and EYEDIAP.

\section*{Acknowledgement}
This work was supported by the National Key R$\&$D Program of China (2016YFB 1001001), the National Natural Science Foundation of China (61976170, 91648121, 61573280), and Tencent Robotics X Lab Rhino-Bird Joint Research Program (201902, 201903). (Portions of) the research in this paper used the EYEDIAP dataset made available by the Idiap Research Institute, Martigny, Switzerland.

\end{document}